\documentclass[sigconf]{acmart}

\usepackage{booktabs} 
\usepackage{graphicx}
\usepackage{amsmath,amssymb} 
\usepackage[utf8]{inputenc}
\usepackage{enumitem}
\usepackage{subcaption}
\usepackage[sort&compress,capitalize,noabbrev,nameinlink]{cleveref}
\creflabelformat{equation}{#2#1#3}
\setcopyright{rightsretained}

\acmDOI{}

\acmISBN{}

\acmConference[ACM Computer Science in Cars Symposium]{ACM Computer Science in Cars Symposium}{2018}{Munich, Germany}
\acmYear{2018}

\acmSubmissionID{29}

\begin{document}
\title{Dense Scene Flow from Stereo Disparity and Optical Flow}
\subtitle{Extended Abstract}

\author{René Schuster}

\author{Oliver Wasenmüller}

\author{Didier Stricker}
\affiliation{%
  \institution{DFKI - German Research Center for Artificial Intelligence}
}
\email{firstname.lastname@dfki.de}

\renewcommand{\shortauthors}{R. Schuster et al.}

\begin{teaserfigure}
	\begin{center}
		\begin{subfigure}[c]{0.32\textwidth}
			\centering
			\includegraphics[width=1\textwidth]{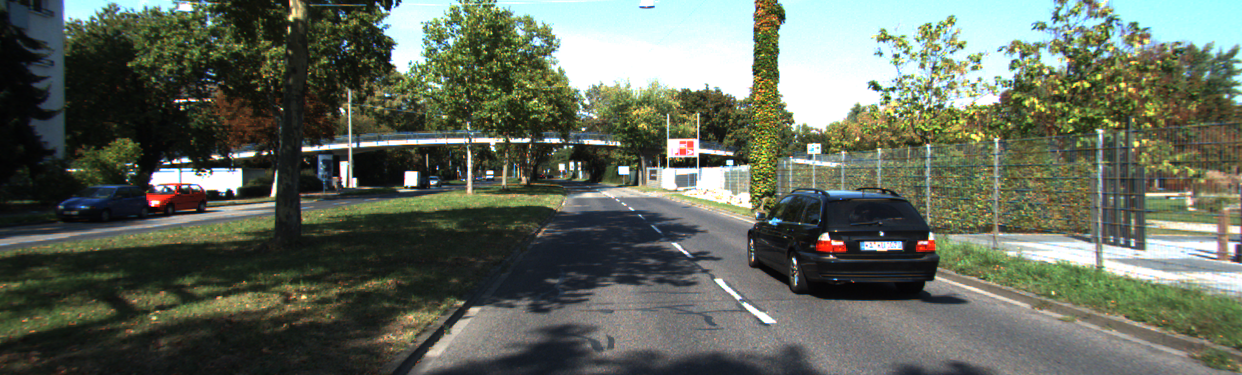}
			\subcaption{Reference image at time $t$.}
			\label{fig:title:img}
		\end{subfigure}
		\begin{subfigure}[c]{0.32\textwidth}
			\centering
			\includegraphics[width=1\textwidth]{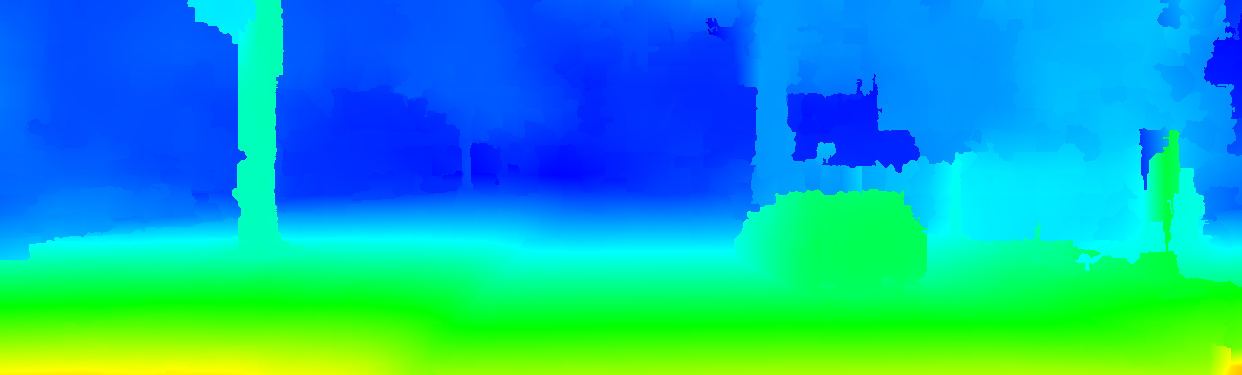}
			\subcaption{Estimated disparity at time $t+1$.}
			\label{fig:title:inputdisp1}
		\end{subfigure}
		\begin{subfigure}[c]{0.32\textwidth}
			\centering
			\includegraphics[width=1\textwidth]{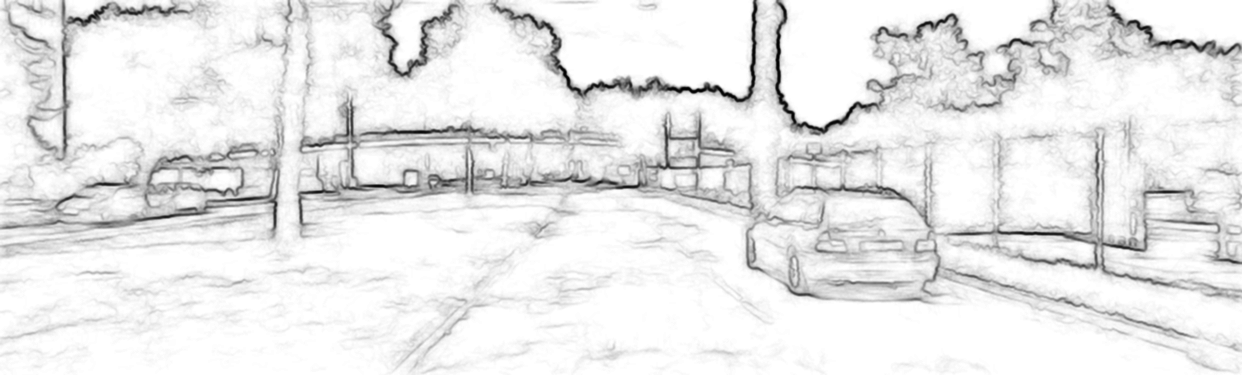}
			\subcaption{Edges used for interpolation.}
			\label{fig:title:edges}
		\end{subfigure}
		\begin{subfigure}[c]{0.32\textwidth}
			\centering
			\includegraphics[width=1\textwidth]{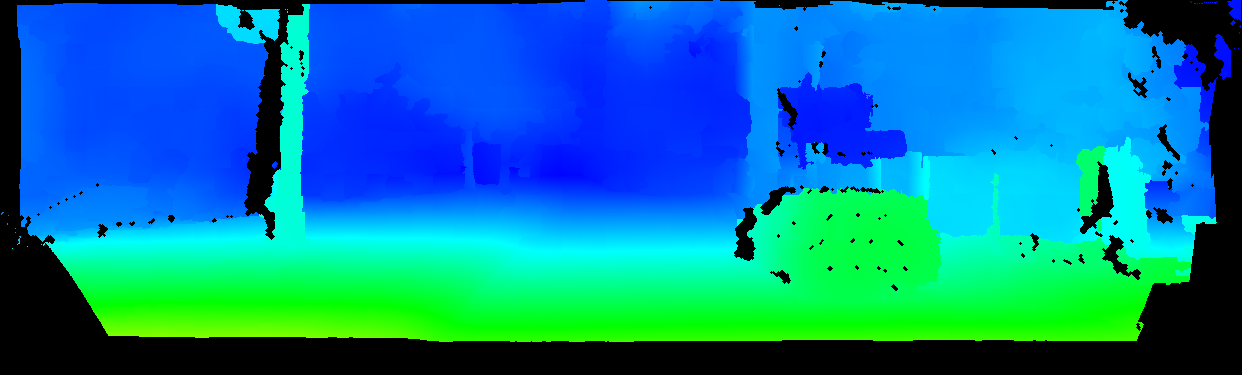}
			\subcaption{Sparse disparity at time $t$.}
			\label{fig:title:sparsedisp0}
		\end{subfigure}
		\begin{subfigure}[c]{0.32\textwidth}
			\centering
			\includegraphics[width=1\textwidth]{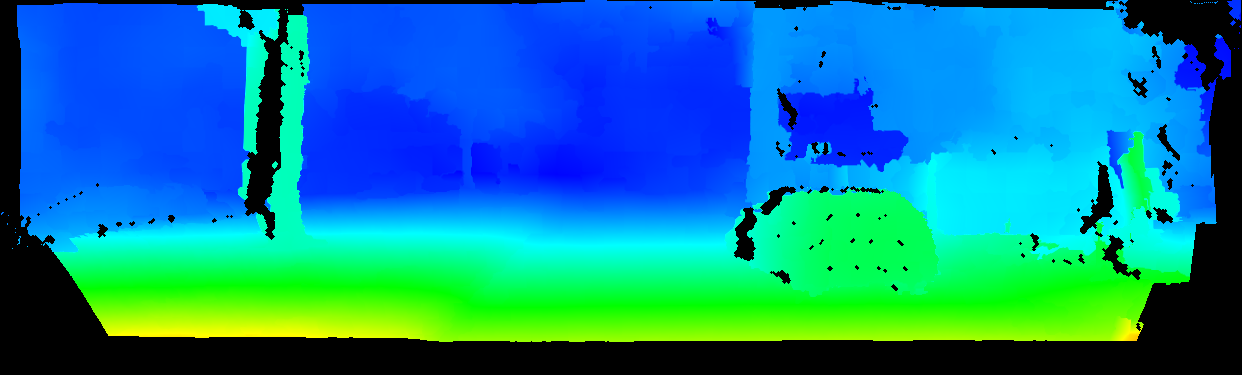}
			\subcaption{Warped, sparse disparity of (\subref{fig:title:inputdisp1}).}
			\label{fig:title:sparsedisp1}
		\end{subfigure}
		\begin{subfigure}[c]{0.32\textwidth}
			\centering
			\includegraphics[width=1\textwidth]{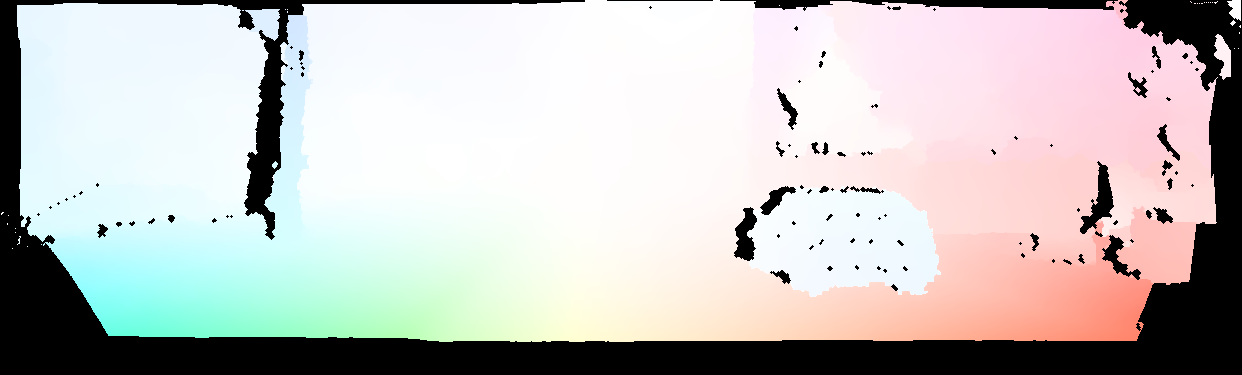}
			\subcaption{Sparse optical flow from $t$ to $t+1$.}
			\label{fig:title:sparseflow}
		\end{subfigure}
		\begin{subfigure}[c]{0.32\textwidth}
			\centering
			\includegraphics[width=1\textwidth]{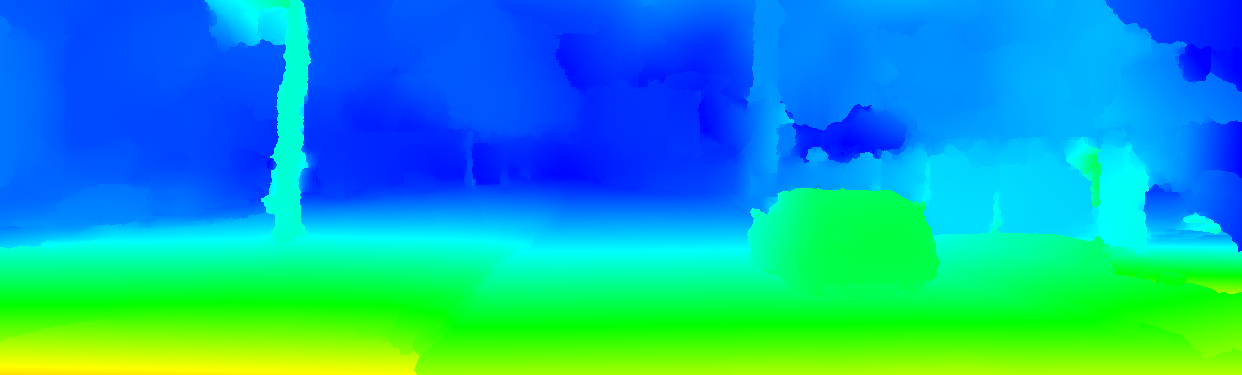}
			\subcaption{Interpolated disparity at time $t$.}
			\label{fig:title:densedisp0}
		\end{subfigure}
		\begin{subfigure}[c]{0.32\textwidth}
			\centering
			\includegraphics[width=1\textwidth]{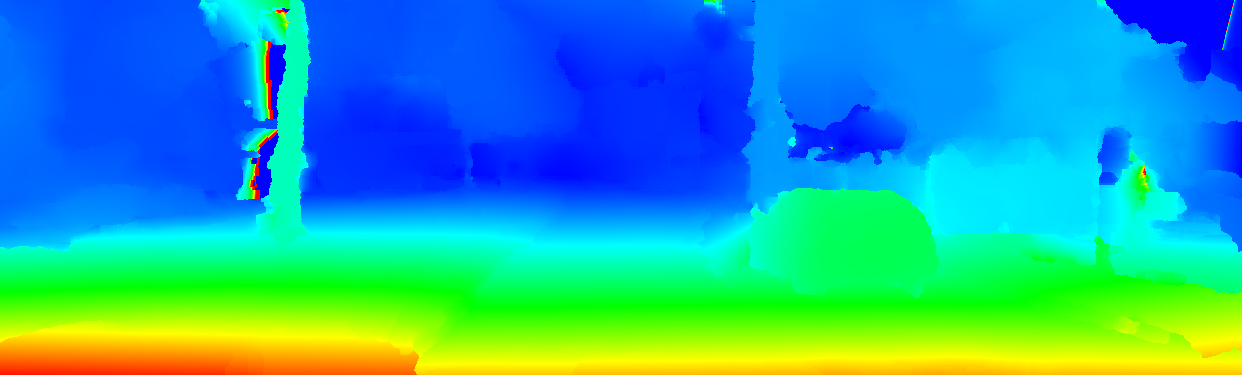}
			\subcaption{Interpolated disparity map of (\subref{fig:title:sparsedisp1}).}
			\label{fig:title:densedisp1}
		\end{subfigure}
		\begin{subfigure}[c]{0.32\textwidth}
			\centering
			\includegraphics[width=1\textwidth]{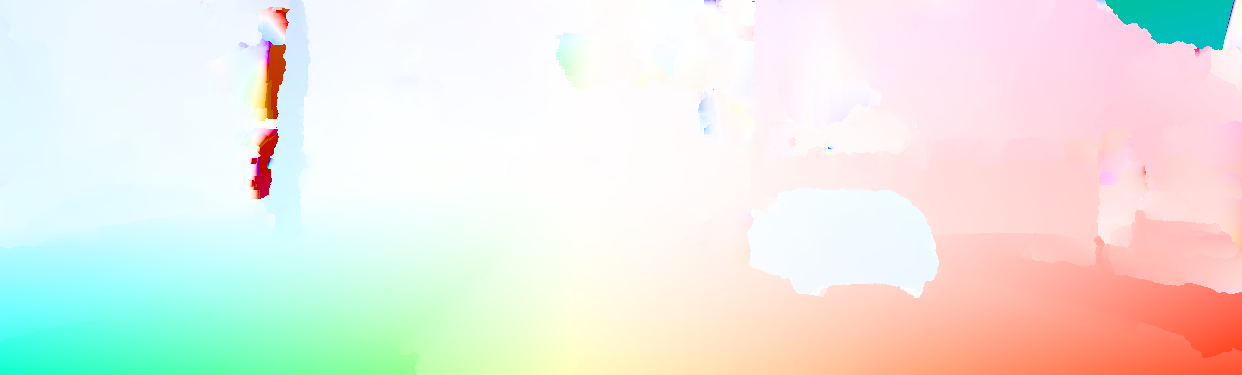}
			\subcaption{Interpolated optical flow.}
			\label{fig:title:denseflow}
		\end{subfigure}
		\caption{Optical flow and stereo disparity are combined to sparse scene flow by warping. Edge-aware interpolation is used to reconstruct a dense scene flow field.}
		\label{fig:title}
	\end{center}
\end{teaserfigure}

\begin{abstract}
Scene flow describes 3D motion in a 3D scene. It can either be modeled as a single task, or it can be reconstructed from the auxiliary tasks of stereo depth and optical flow estimation. While the second method can achieve real-time performance by using real-time auxiliary methods, it will typically produce non-dense results. In this representation of a basic combination approach for scene flow estimation, we will tackle the problem of non-density by interpolation.
\end{abstract}

%
%



\settopmatter{printacmref=false,printccs=false}

\maketitle

\section{Introduction}
\begin{figure}[t]
	\centering
	\includegraphics[width=0.85\columnwidth]{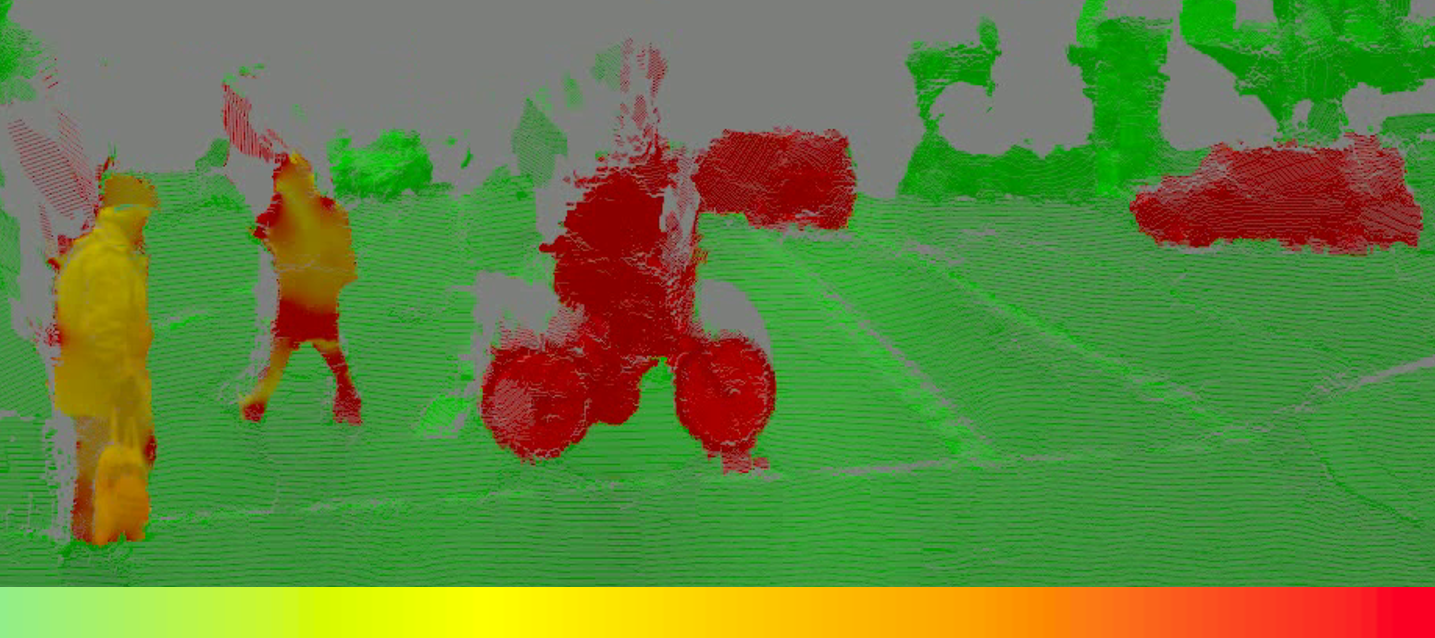}
	\caption{Point cloud visualization of dense scene flow result. Color indicates velocity (green: slow, red: fast).}
	\label{fig:pointcloud}
\end{figure}

The problem of scene flow estimation in computer vision is the reconstruction of 3D geometry and 3D motion based on a sequence of stereo images (see \cref{fig:pointcloud}). While some papers argue that a joint estimation of geometry and motion will yield more consistent results, this paper estimates scene flow by the combination of stereo disparity and optical flow as it was already done in \cite{schuster2018combining}. The sub-tasks are considered computational less expensive and the combination itself is negligible in terms of run-time. Together with a considerably fast and accurate interpolation it is even possible to reconstruct dense scene flow from stereo disparity and optical flow (see \cref{fig:title}).

This is an extension of the already published work in \cite{schuster2018combining}. We will use the interpolation of SceneFlowFields (SFF) \cite{schuster2018sceneflowfields} to reconstruct dense scene flow from the combination of stereo disparity and optical flow.
Relevant related work can be found in the original paper \cite{schuster2018combining}.

\section{Dense Re-Combination}
\paragraph*{\textbf{Sparse Combination.}}
Since scene flow is the description of 3D geometry and 3D motion for every pixel of an image, it can be represented by two vector fields consisting of the 3D position and the 3D displacement in world space. An alternative representation can be given in image space by optical flow, disparity, and disparity change. Given the camera intrinsics and extrinsics, these two representations are equivalent and can be transformed into each other. Consequently, direct computation of optical flow and disparity solves two sub-tasks of scene flow estimation. What is missing is the change of disparity $\Delta d_t^{t+1}$ that together with the disparity $d_t$ yields the disparity at the next time step $d_t^{t+1}=d_t + \Delta d_t^{t+1}$ for each pixel of the reference time step. However, a disparity map $d_{t+1}$ with reference to the next time step can be computed directly and together with optical flow which relates corresponding pixels between both time steps, this disparity map can be warped to the reference frame:
\begin{equation}
	d_t^{t+1}(x,y) = d_{t+1}(x+u,y+v).
\end{equation}
$(u,v)^T$ are the optical flow components at pixel $(x,y)^T$. Thus, we can reconstruct scene flow from optical flow and two disparity maps. Bi-linear interpolation is used to warp disparity values from sub-pixel positions. The only problem is that the reconstruction will fail if the optical flow leaves the image boundaries or where the scene is occluded in the next time step. As a result, the reconstruction approach will produce a non-dense scene flow field (cf. \cref{fig:title:sparsedisp0,fig:title:sparsedisp1,fig:title:sparseflow}).

\begin{table}[t]
	\centering
	\caption{Comparison of different interpolation schemes. Percentage of outliers on KITTI training data.}
	\label{tab:interpolation}
	\begin{tabular}{c c c c c c}
	\toprule
	Interpolation & D1 & D2 & Fl & SF & Density \\
	\midrule
	\textit{sparse} & 4.4 & 8.3 & 9.0 & 12.7 & 84.23 \% \\
	\midrule
	\textit{kitti} & 11.0 & 17.5 & 19.9 & 23.8 & 100.0 \% \\
	\textit{full} & 6.5 & \textbf{12.6} & 16.3 & \textbf{19.6} & 100.0 \% \\
	\textit{motion} & \textbf{4.9} & 13.2 & 16.7 & 20.4 & 100.0 \% \\
	\textit{disp-affine} & \textbf{4.9} & 13.2 & \textbf{15.5} & 20.8 & 100.0 \% \\
	\textit{disp-plane} & \textbf{4.9} & 13.5 & \textbf{15.5} & 21.2 & 100.0 \% \\
	\bottomrule
	\end{tabular}
\end{table}

\paragraph*{\textbf{Interpolation.}}
The recently published SceneFlowFields \cite{schuster2018sceneflowfields} estimates scene flow by sparse-to-dense interpolation. We can utilize the edge-aware interpolation algorithm to reconstruct dense scene flow from the sparse combination of disparity and optical flow. SFF interpolates scene flow in two steps. First, local planes are estimated based on the known scene flow to interpolate geometry. Secondly, local affine transformations are estimated to describe the 3D motion.

Depending on the auxiliary methods that are used for optical flow and disparity estimation, the results for these tasks are already dense. Only the warped disparity map $d_t^{t+1}$ is non-dense. That leaves several options for the interpolation which we all compare in \cref{tab:interpolation}:
\begin{itemize}[topsep=0.5em,leftmargin=*,labelindent=1em]
	\item Using the default interpolation algorithm of the KITTI submission system (\textit{kitti}).
	\item Interpolating all sub-tasks where $d_t^{t+1}$ has gaps (\textit{full}).
	\item Interpolating the 3D motion only (flow + disparity change) according to local affine 3D transformation models (\textit{motion}).
	\item Using the affine 3D transformations to interpolate the warped disparity only (\textit{disp-affine}).
	\item Only interpolating the warped disparity map using a local plane model (\textit{disp-plane}).
\end{itemize} 

\paragraph*{\textbf{Results.}}
Remarkable about the different concepts of interpolation is that the joint interpolation (an example result of this variant is given in \cref{fig:title:densedisp0,fig:title:densedisp1,fig:title:denseflow})  produces overall the best scene flow estimate, though the different sub-results are less accurate than for some other interpolation strategies (see \cref{tab:interpolation}). That supports the general paradigm that scene flow should be estimated jointly.
Further, we want to highlight that the sparse combination results are very accurate. With the steady improvement of methods for the auxiliary tasks, scene flow estimation by re-combination gets better also. In this paper, we have used SPS-stereo \cite{yamaguchi2014efficient} for disparity estimation and FlowFields++ \cite{schuster2018ffpp} for the optical flow tasks. Both are ranked higher than the respective auxiliary methods (SGM \cite{hirschmuller2008SGM} and FF+ \cite{bailer2017optical}) that have been used in the original paper \cite{schuster2018combining}.
Due to that and because the interpolation algorithm we use in this paper is more sophisticated, our dense scene flow estimate from stereo disparity and optical flow is ranked higher in the official public KITTI Scene Flow Benchmark \cite{menze2015object} than the original submission as shown in \cref{tab:results}.

\paragraph*{\textbf{Run-time.}}
The run-time of our approach consists of 29 s for FlowFields++ \cite{schuster2018ffpp}, 2 s for each disparity map computed with SPS-st \cite{yamaguchi2014efficient}, and 3 s for dense interpolation with SceneFlowFields \cite{schuster2018sceneflowfields}.

\begin{table}[t]
	\centering
	\caption{Results according to KITTI Scene Flow Benchmark \cite{menze2015object} in percentage of outliers.}
	\label{tab:results}
	\begin{tabular}{c c c c c c}
	\toprule
	Method & D1 & D2 & Fl & SF & Run-time \\
	\midrule
	ISF \cite{behl2017bounding} & 4.5 & \textbf{6.0} & \textbf{6.2} & \textbf{8.1} & 600 s \\
	PRSM \cite{vogel2015PRSM} & \textbf{4.3} & 6.7 & 6.7 & 9.0 & 300 s \\
	OSF+TC \cite{neoral2017object} & 5.0 & 6.8 & 7.0 & 9.2 & 3000 s \\
	OSF18 \cite{menze2018osf} & 5.3 & 7.1 & 7.4 & 9.7 & 390 s \\
	SSF \cite{ren2017cascaded} & 4.4 & 7.0 & 7.1 & 10.1 & 300 s \\
	OSF \cite{menze2015object} & 5.8 & 7.8 & 7.8 & 10.2 & 3000 s \\
	FSF+MS \cite{taniai2017fsf} & 6.7 & 9.9 & 11.3 & 15.0 & 2.7 s \\
	CSF \cite{lv2016CSF} & 6.0 & 10.1 & 13.0 & 15.7 & 80 s \\
	SFF \cite{schuster2018sceneflowfields} & 6.6 & 10.7 & 12.9 & 15.8 & 65 s \\
	PRSF \cite{vogel2013PRSF} & 6.2 & 12.7 & 13.8 & 16.4 & 150 s \\
	\textbf{Ours} & 6.6 & 14.4 & 16.6 & 20.7 & 36 s \\
	SGM+SF \cite{hirschmuller2008SGM,hornacek2014sphereflow} & 6.8 & 15.6 & 21.7 & 25.0 & 2700 s \\
	PCOF+LDOF \cite{derome2016prediction} & 8.5 & 21.0 & 18.3 & 29.3 & 50 s \\
	SGM+FF+ \cite{schuster2018combining} & 13.4 & 27.8 & 22.8 & 33.6 & 29 s \\
	SGM+C+NL \cite{hirschmuller2008SGM,sun2014quantitative} & 6.8 & 28.25 & 35.6 & 40.3 & 270 s \\
	SGM+LDOF \cite{hirschmuller2008SGM,brox2011large} & 6.8 & 28.6 & 39.3 & 43.7 & 86 s \\
	\bottomrule
	\end{tabular}
\end{table}

\section{Conclusion}
Scene flow estimation by combination of stereo disparity and optical flow is as fast as the auxiliary methods are. The sparse and accurate results can be interpolated to a dense scene flow field with competitive performance. Improvements in stereo algorithms, optical flow estimation, and scene flow interpolation will directly improve the combination approach as presented in this work.

\bibliographystyle{ACM-Reference-Format}
\bibliography{bib}

\end{document}